# CARE: A QLoRA-Fine Tuned Multi-Domain Chatbot with Fast Learning on Minimal Hardware

**Ankit Dutta**
Department of Geology and Geophysics
Indian Institute of Technology, Kharagpur
ankitdutta@kgpian.iitkgp.ac.in

**Nabarup Ghosh**
Department of Geology and Geophysics
Indian Institute of Technology, Kharagpur
nabarupghosh@kgpian.iitkgp.ac.in

**Ankush Chatterjee**[*]
Department of Electrical Engineering
Indian Institute of Technology, Kharagpur
ankushch707@kgpian.iitkgp.ac.in

## Abstract

Large Language models have demonstrated excellent domain-specific question-answering capabilities when finetuned with a particular dataset of that specific domain. However, fine-tuning the models requires a significant amount of training time and a considerable amount of hardware. In this work, we propose CARE (Customer Assistance and Response Engine), a lightweight model made by fine-tuning Phi3.5-mini on very minimal hardware and data, designed to handle queries primarily across three domains: telecommunications support, medical support, and banking support. For telecommunications and banking, the chatbot addresses issues and problems faced by customers regularly in the above-mentioned domains. In the medical domain, CARE provides preliminary support by offering basic diagnoses and medical suggestions that a user might take before consulting a healthcare professional. Since CARE is built on Phi3.5-mini, it can be used even on mobile devices, increasing its usability. Our research also shows that CARE performs relatively well on various medical benchmarks, indicating that it can be used to make basic medical suggestions.

## 1 Introduction

The application of LLMs in industry-specific domains has gained considerable attention in recent years. LLM models are finetuned on a specific domain with a particular dataset and then used for a variety of purposes. Several previous works have explored fine-tuning LLMs for domain-specific applications to enhance their performance in specialized tasks.

BioBERT is a domain-specific adaptation of BERT, fine-tuned on PubMed abstracts and full-text biomedical literature. This fine-tuning enables BioBERT to outperform general-purpose models on biomedical text mining tasks such as named entity recognition (NER), relation extraction, and question answering [1].

FinBERT is a BERT-based model fine-tuned on financial documents, reports, and news to enhance its understanding of finance-related texts. It has been used for sentiment analysis, risk assessment, and fraud detection, significantly improving upon non-domain-specific models [2].

However, these models solve the problems of a single domain. If anyone wants to get a model proficient in various domains and yet at the same time be lightweight, then previously they must have used a text classifier which could classify the domain of the text and then direct it to the domain specific model based on the classification. This induces chances of error and at the same time increases memory requirements significantly.

Hence, In this work, we introduce CARE, a lightweight model based on Phi-3.5-mini, fine-tuned using QLoRA (Quantized Low Rank Adaptation) to enable efficient multi-domain conversational assistance. Unlike traditional fine-tuning approaches that update all model parameters,

---
[*]Corresponding Author

QLoRA fine-tunes only a subset of the model's weights, reducing memory usage and computational costs [3]. This model is proficient in solving multi-domain problems and yet it keeps the memory requirement of the user very low, reducing costs.

CARE [2] is designed to handle queries spanning three key industry domains: healthcare, banking, and telecommunications. By leveraging QLoRA, we ensure the model maintains high adaptability without excessive computational costs [3][4]. We evaluate CARE across domain-specific case studies to assess its performance and benchmark it against existing chatbot solutions. Our findings indicate that CARE can be used as a model to solve multi-domain problems due to its good performance in the medical benchmarks and satisfactory response generation for banking support and telecommunications support queries.

## 2 Methodology

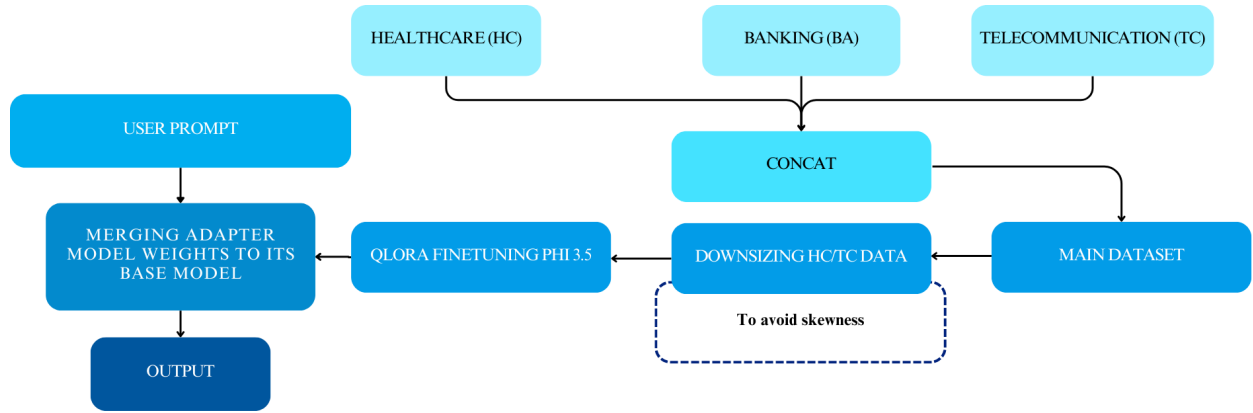

Figure 1: Flowchart illustrating the dataset preparation and training process for CARE

### 2.1 Datasets

Figure 1 shows a flowchart illustrating the dataset preparation and training process for CARE. The model was built upon three distinct datasets, each representing a different domain: Medical Queries Support, Banking Queries Support, and Telecommunication Queries Support. These datasets were obtained from publicly available repositories in Kaggle [5] and Hugging Face [6][7].

To construct a comprehensive training corpus, the datasets were initially merged into a unified dataset. However, a significant class imbalance was observed, as the medical dataset contained approximately 21 million samples [7], whereas the Telecommunications dataset [6] comprised around 26,000 samples, while the Banking dataset was considerably smaller than both. This imbalance adversely affected the fine-tuning process, leading to extended training durations and suboptimal model performance, particularly on Banking-related queries.

To mitigate these challenges, a down sampling strategy was implemented for the Medical and Telecommunications datasets, aligning their sizes with that of the Banking dataset. This adjustment facilitated a more balanced learning process, ensuring that the model could generalize effectively across all three domains without being disproportionately influenced by any single dataset [8].

Finally, the down-sampled Medical and Banking dataset was merged with the Banking dataset to create the final training dataset. Each row of the dataset contained three values: the query, the response, and the class/domain to which the query belongs.

### 2.2 Model Training

In this study, we fine-tuned the base model using QLoRA and PEFT (Parameter-Efficient Fine-Tuning) to develop CARE. The fine-tuning process utilized Supervised Fine-Tuning (SFT), followed by merging the adapter weights with the base model to create a fully integrated system [3; 8]. We used Microsoft's Phi-3.5 4K Mini Instruct, a lightweight yet powerful instruction-tuned model capable of handling extended contexts up to 4K tokens as our base model [9].

To enable efficient fine-tuning with limited computational resources, we leveraged QLoRA (Quantized Low-Rank Adaptation), which enables training a large model with significantly reduced memory consumption [3; 4]. The following key python libraries were used:

- Transformers: For defining the model and tokenizer.
- Peft: For creating LoRA adapters [10].

[2]https://huggingface.co/CARIFY-AI/CARE

- Bitsandbytes: For 4-bit quantization to reduce memory usage [10].
- Trl: For supervised fine-tuning and chat format setup [10].

We then initialized PEFT to introduce LoRA (Low-Rank Adaptation) layers into the model, which enables efficient adaptation of pre-trained models by introducing small trainable parameters while keeping most of the model weights frozen. This method significantly reduces computational overhead and memory usage during fine-tuning [10]. The LoRA configuration was carefully designed to optimize learning efficiency while ensuring robustness in adapting to new data distributions. The configuration parameters included a lora-alpha value of 32, which scales the updated weights appropriately, and a lora-dropout rate of 0.05 to prevent overfitting by introducing stochastic regularization . Additionally, the rank r = 8 was chosen to balance expressiveness and efficiency in low-rank matrix updates [3; 4]. The bias term was set to "none," ensuring that no additional bias parameters were learned, and the task-type was set to "CAUSAL-LM," which specifies its usage for autoregressive language modeling tasks [10].

The Phi-3.5 4K Mini Instruct model was then modified to support LoRA-based fine-tuning by integrating these configurations within the PEFT framework, allowing the model to be fine-tuned with minimal hardware requirements while maintaining its generative capabilities [9].

### 2.3 Supervised Fine Tuning

The training data was loaded from a preprocessed dataset containing multi-turn dialogue interactions for CARE. The dataset was tokenized and structured for SFT, ensuring compatibility with LoRA-based training [8]. The model was fine-tuned using the QLoRA method, where only the adapter layers were updated while keeping the base model weights frozen [3; 4].

After fine-tuning, the adapter weights were merged with the base model using peft library [10]. This process effectively integrated the fine-tuned parameters into the main model.

### 2.4 Training Details

The model was trained for 118 global steps, achieving a final training loss of 1.7327. Figure 2 and Figure 3 shows the loss value and the learning rate for every global step during the finetuning process for CARE. The training process was conducted on a Linux-based system (Kernel version 6.6.56+) using CPython 3.10.12.

The training script was executed in a Kaggle Notebook environment, utilizing a Tesla P100-PCIE-16GB GPU with 3,584 CUDA cores, based on the Pascal architecture, and a CUDA version of 12.6 [11]. The system had a total available memory of 33.66 GB and 4 logical CPU cores as described in Table 1.

Table 1: System Information

| System Info | Details |
| --- | --- |
| OS | Linux-6.6.56+-x86_64 |
| Python Version | CPython 3.10.12 |
| CPU Count | 2 |
| Logical CPU Count | 4 |
| GPU | Tesla P100-PCIE-16GB |
| GPU Count | 1 |
| Total Disk Space (bytes) | 8.66 TB |
| Used Disk Space (bytes) | 6.63 TB |
| Total Memory (bytes) | 33.66 GB |
| CUDA Version | 12.6 |
| GPU Architecture | Pascal |
| CUDA Cores | 3584 |
| GPU Memory (bytes) | 17.18 GB |

The total floating-point operations (FLOPs) during training were computed as $3.75 \times 10^{15}$. The training process spanned 608.84 seconds, achieving a throughput of 1.552 samples per second and 0.194 steps per second. The mean token accuracy recorded during training was 69.54%, and the global gradient norm was 0.3081, indicating stable gradient updates. The Adam optimizer was used for the training process [12]. A detailed summary of these training metrics is provided in Table 2.

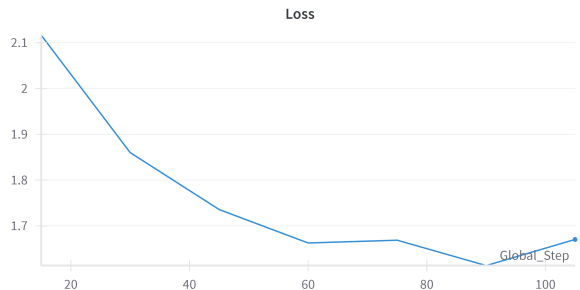

Figure 2 : This figure shows the loss value for every global step during the finetuning process

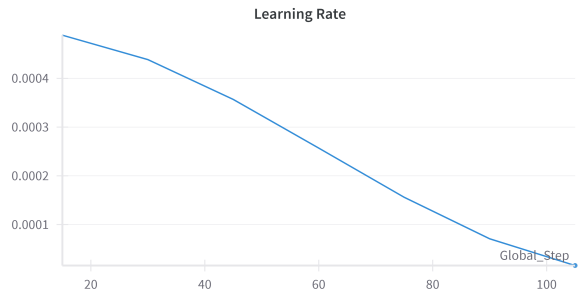

Figure 3: This figure shows the learning rate with every global step during finetuning

Table 2 : Training Metrics

| Metric | Value |
| --- | --- |
| Epoch | 0.99899 |
| Loss | 1.6703 |
| Steps per second | 0.194 |
| Samples per second | 1.552 |
| Total FLOPs | 3.75e15 |
| Runtime (s) | 608.8373 |
| Gradient Norm | 0.3081 |
| Global Step | 118 |
| Learning Rate | 1.587e-5 |
| Train Loss | 1.7327 |
| Mean Token Accuracy | 0.6954 |
| Step | 7 |
| Timestamp | 1741203641.65 |
| Wandb Runtime | 626 |
| Runtime (alt) | 618.2498 |

## 2.5 Model Evaluation

To assess the performance of CARE, the model was benchmarked on publicly available datasets in the medical domain, with results presented in Table 3 and visualized through a histogram in Figure 4. Due to the lack of standardized datasets and resources, similar evaluations could not be conducted for the banking and telecom domains.

### 2.5.1 Medical Domain Performance

On the MedMCQA dataset, CARE achieved a score of 54.3, surpassing Llama-3.2-3b-Instruct (50.9) and significantly outperforming Gemma-2-2b-it (43.0) and DeepSeek-R1-Distill-Qwen2.5 (32.1) [13]. This indicates CARE's superior ability to understand medical multiple-choice questions.

CARE also demonstrated high competency in MMLU Anatomy, achieving 63.7, a notable improvement over Llama-3.2-3b-Instruct (59.3), while vastly outperforming Gemma-2-2b-it (51.9) and DeepSeek-R1-Distill-Qwen2.5 (29.6).[14]

In MMLU Clinical Knowledge, CARE obtained the highest accuracy (78.1), showcasing its strength in medical reasoning and problem-solving. Similarly, it performed best in MMLU College Biology (84.7), MMLU College Medicine (67.6), and MMLU Medical Genetics (76.0), consistently exceeding the performance of other models.[14]

For PubMedQA, a dataset designed to evaluate models' understanding of biomedical literature, CARE achieved 76.4, demonstrating its robust comprehension of scientific texts. This score surpasses Llama-3.2-3b-Instruct (69.8) and significantly outperforms DeepSeek-R1-Distill-Qwen2.5 (60.0) [15].

Overall, the results indicate that CARE consistently achieves state-of-the-art performance in medical and clinical knowledge benchmarks, highlighting its suitability for healthcare and biomedical applications.

### 2.5.2 Other Domain Performance

Despite its strong performance in the medical domain, CARE could not be systematically evaluated for banking support and telecommunications support due to the scarcity of standardized datasets specific to these domains.

The absence of these benchmarks presents a limitation in quantitatively assessing CARE's generalization to these domains. Future work should focus on:

- Developing domain-specific datasets for banking and telecommunications support.

- Exploring alternative evaluation methodologies, such as expert-driven assessments and real-world deployment feedback.

- Making synthetic datasets to test domain-specific question-answering tasks.

A structured evaluation of CARE across all three domains will enable a more comprehensive understanding of its effectiveness and generalization capabilities in specialized industry settings.

## 2.6 Model Comparison

A comparative analysis of CARE with other prominent models—Llama-3.2-3b-Instruct, Gemma-2-2b-it, and DeepSeek-R1-Distill-Qwen2.5—reveals some very key distinctions in performance. Better Performance than Llama-3.2-3b-Instruct: CARE surpasses Llama-3.2-3b-Instruct in almost every benchmark, with notable improvements in MMLU Clinical Knowledge (+15.5 points), MMLU College Biology (+13.9 points), and MMLU College Medicine (+9.2 points) [13; 14; 15].

- **High Medical Knowledge Reasoning**: CARE outperforms all other models across every dataset, demonstrating its enhanced capability in answering complex medical questions with high accuracy [15].

- **Better performance than Gemma-2-2b-it**: While Gemma-2-2b-it shows decent performance, CARE performs significantly better than it across all benchmarks, highlighting its adaptability to specialized domains.

- **Strong Comprehension of Scientific Literature**: CARE achieved a very high score on PubMedQA (76.4), proving its effectiveness in handling biomedical texts, a critical area for AI applications in research and healthcare [15].

This comparative analysis shows that CARE has performed better than most of the other models with a size in the same or close range (1.5 B - 3 B), which shows that CARE can be used as a medical support. The model also generates satisfactory respones on banking support and telecommunication support domains. Further information can be obtained regarding the performance of various models by contacting the authors of this paper.

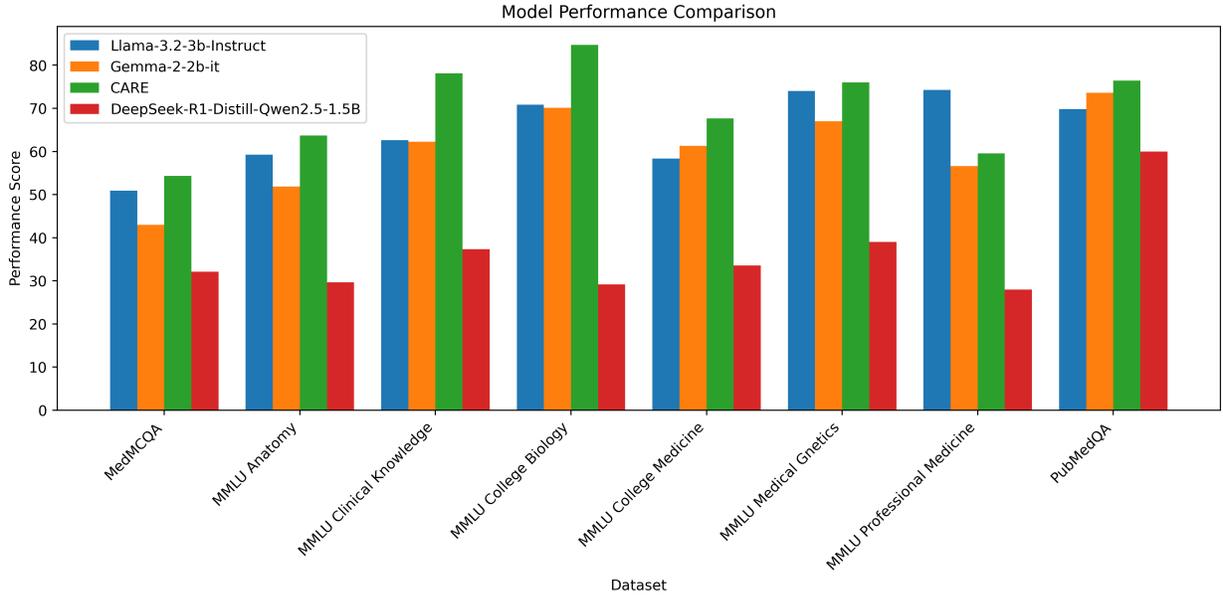

Figure 4: Histogram comparison of medical model performance across benchmarks

Table 3: Model performance on various medical benchmarks

| Dataset | Llama-3.2-3b-Instruct | Gemma-2-2b-it | CARE | DeepSeek-R1-Distill-Qwen2.5-1.5B |
|---|---|---|---|---|
| MedMCQA | 50.9 | 43.0 | **54.3** | 32.1 |
| MMLU Anatomy | 59.3 | 51.9 | **63.7** | 29.6 |
| MMLU Clinical Knowledge | 62.6 | 62.3 | **78.1** | 37.4 |
| MMLU College Biology | 70.8 | 70.1 | **84.7** | 29.2 |
| MMLU College Medicine | 58.4 | 61.3 | **67.6** | 33.5 |
| MMLU Medical Genetics | 74.0 | 67.0 | **76.0** | 39.0 |
| MMLU Professional Medicine | **74.3** | 56.6 | 59.5 | 27.9 |
| PubMedQA | 69.8 | 73.6 | **76.4** | 60.0 |

## 3 Conclusion

In this study, we presented CARE, a chatbot fine-tuned using QLoRA on the Phi-3.5 4K-Mini-Instruct model for multi-domain conversational assistance. By leveraging parameter-efficient fine-tuning techniques, we significantly reduced computational overhead while maintaining robust performance across specialized domains [3]. Our evaluation in the medical domain showcased CARE's superior performance compared to existing models, underscoring its ability to comprehend complex medical queries.

Despite these promising results, the lack of standardized evaluation benchmarks for banking and telecommunications domains limited our ability to assess CARE's performance comprehensively. This limitation highlights the need for domain-specific datasets to enable rigorous benchmarking in real-world applications. Furthermore, integrating reinforcement learning and improving the chatbot's knowledge base across additional industries could enhance its adaptability and performance [8].

## 4 Future Scope

Several avenues for future research and development emerge from our findings:

- **Development of Domain-Specific Benchmarks** – Creating standardized datasets for banking and telecommunications question-answering tasks will facilitate systematic evaluations and comparisons.

- **Expanding CARE's Knowledge Base** – Incorporating additional industry-specific data, including legal and financial domains, can enhance CARE's versatility[8].

- **Integrating Reinforcement Learning** – Employing Reinforcement Learning with Human Feedback (RLHF) can improve CARE's contextual understanding and response generation[8].

By addressing these areas, future work can further enhance CARE's capabilities, making it a robust and efficient AI-powered assistant for diverse industry applications.

# References

[1] Jinhyuk Lee, Wonjin Yoon, Sungdong Kim, Donghyeon Kim, Sunkyu Kim, Chan Ho So, and Jaewoo Kang. Biobert: a pre-trained biomedical language representation model for biomedical text mining. *Bioinformatics*, 36(4):1234–1240, 09 2019. ISSN 1367-4803. doi:10.1093/bioinformatics/btz682. URL https://doi.org/10.1093/bioinformatics/btz682.

[2] Dogu Araci. Finbert: Financial sentiment analysis with pre-trained language models, 2019. URL https://arxiv.org/abs/1908.10063.

[3] Tim Dettmers, Artidoro Pagnoni, Ari Holtzman, and Luke Zettlemoyer. Qlora: Efficient finetuning of quantized llms. In *Advances in Neural Information Processing Systems*, volume 36, pages 10088–10115, 2023. URL https://arxiv.org/abs/2305.14314.

[4] Edward J. Hu, Yelong Shen, Phillip Wallis, Zeyuan Allen-Zhu, Yuanzhi Li, Shean Wang, Lu Wang, and Weizhu Chen. Lora: Low-rank adaptation of large language models. In *International Conference on Learning Representations (ICLR)*, 2022.

[5] Kaggle Datasets. Banking chatbot dataset. https://www.kaggle.com/datasets/manojajj/banking-chatbot.

[6] Hugging Face Datasets. Bitext-telco-llm-chatbot-training-dataset. https://huggingface.co/datasets/bitext/Bitext-telco-llm-chatbot-training-dataset.

[7] Ruslan Magana Vsevolodovna. Ai medical dataset, 2023. URL https://github.com/ruslanmv/ai-medical-chatbot.

[8] Brian Lester, Rami Al-Rfou, and Noah Constant. The power of scale for parameter-efficient prompt tuning, 2021. URL https://arxiv.org/abs/2104.08691.

[9] Marah Abdin, Jyoti Aneja, Hany Awadalla, Ahmed Awadallah, Ammar Ahmad Awan, Nguyen Bach, Amit Bahree, Arash Bakhtiari, Jianmin Bao, Harkirat Behl, Alon Benhaim, Misha Bilenko, Johan Bjorck, Sébastien Bubeck, Martin Cai, Qin Cai, Vishrav Chaudhary, Dong Chen, Dongdong Chen, Weizhu Chen, Yen-Chun Chen, Yi-Ling Chen, Hao Cheng, Parul Chopra, Xiyang Dai, Matthew Dixon, Ronen Eldan, Victor Fragoso, Jianfeng Gao, Mei Gao, Min Gao, Amit Garg, Allie Del Giorno, Abhishek Goswami, Suriya Gunasekar, Emman Haider, Junheng Hao, Russell J. Hewett, Wenxiang Hu, Jamie Huynh, Dan Iter, Sam Ade Jacobs, Mojan Javaheripi, Xin Jin, Nikos Karampatziakis, Piero Kauffmann, Mahoud Khademi, Dongwoo Kim, Young Jin Kim, Lev Kurilenko, James R. Lee, Yin Tat Lee, Yuanzhi Li, Yunsheng Li, Chen Liang, Lars Liden, Xihui Lin, Zeqi Lin, Ce Liu, Liyuan Liu, Mengchen Liu, Weishung Liu, Xiaodong Liu, Chong Luo, Piyush Madan, Ali Mahmoudzadeh, David Majercak, Matt Mazzola, Caio César Teodoro Mendes, Arindam Mitra, Hardik Modi, Anh Nguyen, Brandon Norick, Barun Patra, Daniel Perez-Becker, Thomas Portet, Reid Pryzant, Heyang Qin, Marko Radmilac, Liliang Ren, Gustavo de Rosa, Corby Rosset, Sambudha Roy, Olatunji Ruwase, Olli Saarikivi, Amin Saied, Adil Salim, Michael Santacroce, Shital Shah, Ning Shang, Hiteshi Sharma, Yelong Shen, Swadheen Shukla, Xia Song, Masahiro Tanaka, Andrea Tupini, Praneetha Vaddamanu, Chunyu Wang, Guanhua Wang, Lijuan Wang, Shuohang Wang, Xin Wang, Yu Wang, Rachel Ward, Wen Wen, Philipp Witte, Haiping Wu, Xiaoxia Wu, Michael Wyatt, Bin Xiao, Can Xu, Jiahang Xu, Weijian Xu, Jilong Xue, Sonali Yadav, Fan Yang, Jianwei Yang, Yifan Yang, Ziyi Yang, Donghan Yu, Lu Yuan, Chenruidong Zhang, Cyril Zhang, Jianwen Zhang, Li Lyna Zhang, Yi Zhang, Yue Zhang, Yunan Zhang, and Xiren Zhou. Phi-3 technical report: A highly capable language model locally on your phone, 2024. URL https://arxiv.org/abs/2404.14219.

[10] Sourab Mangrulkar, Sylvain Gugger, Lysandre Debut, Younes Belkada, Sayak Paul, and Benjamin Bossan. Peft: State-of-the-art parameter-efficient fine-tuning methods. https://github.com/huggingface/peft.

[11] Tesla p100-pcie-16gb gpu. https://images.nvidia.com/content/tesla/pdf/nvidia-tesla-p100-datasheet.pdf.

[12] Diederik P. Kingma and Jimmy Ba. Adam: A method for stochastic optimization, 2015. URL https://arxiv.org/abs/1412.6980.

[13] Ankit Pal, Logesh Kumar Umapathi, and Malaikannan Sankarasubbu. Medmcqa : A large-scale multi-subject multi-choice dataset for medical domain question answering, 2022. URL https://arxiv.org/abs/2203.14371.

[14] Dan Hendrycks, Collin Burns, Steven Basart, Andy Zou, Mantas Mazeika, Dawn Song, and Jacob Steinhardt. Measuring massive multitask language understanding. *arXiv preprint*, arXiv:2009.03300, 2020. URL https://arxiv.org/abs/2009.03300.

[15] Qiao Jin, Bhuwan Dhingra, Zhengping Liu, William Cohen, and Xinghua Lu. PubMedQA: A dataset for biomedical research question answering. In Kentaro Inui, Jing Jiang, Vincent Ng, and Xiaojun Wan, editors, *Proceedings of the 2019 Conference on Empirical Methods in Natural Language Processing and the 9th International Joint Conference on Natural Language Processing (EMNLP-IJCNLP)*, pages 2567–2577, Hong Kong, China, November 2019. Association for Computational Linguistics. doi:10.18653/v1/D19-1259. URL https://aclanthology.org/D19-1259/.